\documentclass[conference]{IEEEtran}
\IEEEoverridecommandlockouts

\usepackage{amsmath,amssymb,bm}
\usepackage{graphicx}
\usepackage{booktabs}
\usepackage{array}
\usepackage[dvipsnames]{xcolor} 
\usepackage{cite}
\usepackage{url}
\usepackage{tikz}
\usetikzlibrary{arrows.meta,positioning,fit,calc,shapes.misc}
\usepackage{microtype}
\usepackage{balance}
\usepackage{bm}
\graphicspath{{figures/}{./}}

\definecolor{darkgray}{RGB}{65,65,65}
\definecolor{inputgray}{RGB}{244,244,244}
\definecolor{modelblue}{RGB}{226,236,248}
\definecolor{evidenceorange}{RGB}{255,238,218}
\definecolor{fusionyellow}{RGB}{255,250,224}
\definecolor{auditgreen}{RGB}{229,244,232}

\newcommand{\method}{OAS-MIL}
\newcommand{\xv}{\bm{x}}
\newcommand{\zv}{\bm{z}}
\newcommand{\hv}{\bm{h}}
\newcommand{\qv}{\bm{q}}
\newcommand{\uv}{\bm{u}}
\newcommand{\vv}{\bm{v}}
\newcommand{\bv}{\bm{b}}
\newcommand{\W}{\bm{W}}
\newcommand{\LN}{\operatorname{LN}}
\newcommand{\GELU}{\operatorname{GELU}}
\newcommand{\sigmoid}{\sigma}

\tikzset{
  block/.style={draw=darkgray, rounded corners=2pt, fill=inputgray, thick, align=center, minimum height=7mm, font=\scriptsize, text width=25mm},
  model/.style={draw=darkgray, rounded corners=2pt, fill=modelblue, thick, align=center, minimum height=7mm, font=\scriptsize, text width=28mm},
  evidence/.style={draw=darkgray, rounded corners=2pt, fill=evidenceorange, thick, align=center, minimum height=7mm, font=\scriptsize, text width=31mm},
  fuse/.style={draw=darkgray, rounded corners=2pt, fill=fusionyellow, thick, align=center, minimum height=7mm, font=\scriptsize, text width=30mm},
  audit/.style={draw=darkgray, rounded corners=2pt, fill=auditgreen, thick, align=center, minimum height=7mm, font=\scriptsize, text width=28mm},
  cell/.style={draw=darkgray, rounded corners=1.5pt, fill=white, thick, align=center, minimum width=8mm, minimum height=5mm, font=\scriptsize},
  arrow/.style={-Latex, thick, draw=darkgray},
  bluearrow/.style={-Latex, thick, draw=blue!55!black},
  orangearrow/.style={-Latex, thick, draw=orange!75!black}
}

\newcolumntype{L}[1]{>{\raggedright\arraybackslash}p{#1}}
\newcolumntype{C}[1]{>{\centering\arraybackslash}p{#1}}

\makeatletter
\def\@IEEEauthorblockconfadjspace{-.9em}
\makeatother

\begin{document}

\title{Order-Aware 2.5D Multiple Instance Learning for Preoperative MRI-Based Perineural Invasion Risk Assessment in Intrahepatic Cholangiocarcinoma}

\author{
\IEEEauthorblockN{
Hyunsu Go$^{1}$, Youngung Han$^{1,2}$, Kyeonghun Kim$^{2}$, Jinyong Jun$^{1}$, Junbeom Lee$^{3}$, \\ 
Dohyun Kweon$^{2,4}$, Yului Jeong$^{1}$, Suah Park$^{1}$, Sungha Park$^{1}$, Anna Jung$^{1}$, \\
Woo Kyoung Jeong$^{5}$, Ken Ying-Kai Liao$^{6}$, Hyuk-Jae Lee$^{1}$, Nam-Joon Kim$^{1,\dagger}$
}

\IEEEauthorblockA{\footnotesize
$^{1}$Seoul National University, Seoul, Republic of Korea \quad
$^{2}$OUTTA, Seoul, Republic of Korea
}

\IEEEauthorblockA{\footnotesize
$^{3}$Chung-Ang University, Seoul, Republic of Korea  \quad
$^{4}$ Kyung Hee University, Seoul, Republic of Korea 
}

\IEEEauthorblockA{\footnotesize
$^{5}$Samsung Medical Center, Seoul, Republic of Korea \quad
$^{6}$NVIDIA AI Technology Center, Taipei, Taiwan
}

\IEEEauthorblockA{\footnotesize
$^{\dagger}$Corresponding author: \texttt{knj01@snu.ac.kr}
}
}

\IEEEaftertitletext{\vspace{-.9em}}
\maketitle


\begin{abstract}
Perineural invasion (PNI) is an adverse histopathologic marker in intrahepatic cholangiocarcinoma (ICC), but it is usually confirmed only after resection. Preoperative T2-weighted MRI may provide noninvasive imaging cues predictive of PNI, although labels are available only at the patient level without slice- or voxel-level annotations. We propose Order-Aware Slab Multiple Instance Learning (\method{}), a weakly supervised framework for patient-level PNI prediction. Each tumor-centered MRI crop is represented as an ordered sequence of overlapping 2.5D slabs formed from contiguous axial slices. A shared encoder extracts slab-level features, which are aggregated by a permutation-invariant set-attention branch and a bidirectional sequence-attention branch. Using five-fold label-stratified cross-validation at the patient level, \method{} achieved a mean AUROC of $0.770 \pm 0.077$, outperforming the evaluated volumetric and MIL baselines. These results suggest that axial order provides a useful inductive bias for weakly supervised PNI prediction from MRI.
\end{abstract}


\section{Introduction}
Intrahepatic cholangiocarcinoma (ICC) is an aggressive primary liver malignancy for which surgical resection remains the principal curative-intent treatment. However, postoperative recurrence is common and long-term outcomes remain limited \cite{banales2020cholangiocarcinoma}. Perineural invasion (PNI), defined as tumor involvement of nerves or perineural spaces, is a histopathologic marker of tumor aggressiveness \cite{liebig2009pni}. In ICC, surgical studies have associated PNI with recurrence and worse survival after curative-intent resection \cite{shirai2008pni, zhang2020pni}. 
These findings motivate preoperative PNI prediction as a complementary risk-stratification approach.

Despite its clinical relevance, PNI is difficult to identify preoperatively because the reference standard is microscopic pathologic assessment after surgery, whereas routine MRI does not provide nerve-level or slice-level labels for model training \cite{liebig2009pni}. Previous studies have investigated noninvasive PNI prediction using CT radiomics and clinicoradiological machine-learning models \cite{liu2024ct}. More recently, deep learning approaches have explored PNI prediction from 3D MRI \cite{han2026losa,go2026anatomy,han2026mma}. These studies support image-based PNI risk modeling, but they also highlight a weak-supervision gap: the endpoint is assigned at the patient level, while local imaging evidence is not annotated.

Medical images can encode tumor phenotypes that are relevant to clinical endpoints \cite{lambin2012radiomics,gillies2016radiomics}. For MRI-based PNI prediction, however, a model must aggregate information over a tumor-centered volume without assuming slice-, voxel-, or nerve-level supervision. Multiple instance learning (MIL) is well-suited to this setting because a bag of instances can be supervised by one bag-level label \cite{dietterich1997mil,maron1997mil,carbonneau2018mil}. Attention-based MIL provides a differentiable bag-pooling mechanism \cite{ilse2018attention}, and related weakly supervised aggregation approaches have been widely used in medical image analysis and computational pathology \cite{lu2021clam,li2021dsmil,shao2021transmil}.

A key modeling issue is how to represent volumetric MRI. Fully volumetric encoders preserve 3D context, but their memory footprint and optimization burden are higher than slice-wise processing, especially when clinical cohorts are small \cite{cicek2016unet,milletari2016vnet,tajbakhsh2016cnn}. Slice-wise 2D encoders are efficient but cannot directly model local through-plane continuity. A 2.5D formulation, in which adjacent slices are stacked as channels, offers a practical compromise by retaining local cross-slice information while using efficient 2D-style feature extraction \cite{setio2016pulmonary,roth2015deeporgan}. In tumor-centered MRI, such slabs are not exchangeable observations: their axial order encodes anatomical continuity. Permutation-invariant MIL may therefore underuse through-plane patterns available in the ordered slab sequence.

This paper proposes Order-Aware Slab MIL (\method{}) for weakly supervised prediction of PNI status from preoperative T2-weighted MRI. The model retains a set-attention branch and adds a sequence-attention branch that reads the slab sequence with bidirectional context before patient-level pooling. The contributions are as follows:
\begin{itemize}
  \item We formulate MRI-based PNI prediction as ordered 2.5D MIL, where each tumor-centered T2-weighted MRI crop is represented as a complete sequence of contiguous slabs under patient-level supervision.
  \item We introduce a sequence-attention branch that models axial order with a bidirectional recurrent reader and produces normalized slab-level aggregation weights.
  \item We evaluate the proposed model against volumetric and MIL baselines using five-fold label-stratified cross-validation at the patient level, with ablations of the slab representation, branch aggregation, and sequence reader.
\end{itemize}

\begin{figure*}[t] 
\centering

\usetikzlibrary{positioning, arrows.meta, fit, calc, decorations.pathreplacing, backgrounds}

\def\sliceW{1.45cm}
\def\slabBagW{1.45cm}
\def\flowGap{11mm}
\def\encSideGap{\flowGap}
\def\featXOffset{1.425}
\def\branchXOffset{2.35}
\def\headXOffset{1.50}

\def\figXShift{-5.0mm}
\definecolor{silver}{RGB}{192,192,192}
\def\opBlockW{28mm}
\def\opBlockH{7.5mm}
\def\opBlockLW{0.7pt}

\def\featBlockW{4.5mm}
\def\featBlockH{2.2mm}
\def\featBoxW{7.5mm}

\noindent\makebox[\linewidth][c]{%
\hspace*{\figXShift}%
\resizebox{0.98\textwidth}{!}{%
\begin{tikzpicture}[
  node distance=8mm and 9mm,
  figlabel/.style={
    font=\scriptsize,
    text=black
  },
  opblock/.style={
    draw=black,
    rounded corners=2pt,
    line width=\opBlockLW,
    align=center,
    figlabel,
    inner sep=2pt
  },
  figbranch/.style={
    opblock,
    text width=24mm,
    minimum width=\opBlockW,
    minimum height=\opBlockH
  },
  figset/.style={figbranch, fill=modelblue},
  figord/.style={figbranch, fill=evidenceorange},
  fighead/.style={
    opblock,
    fill=silver,
    minimum width=\opBlockH,
    minimum height=\opBlockW
  },
  featblock/.style={
    draw=gray!70,
    fill=silver,
    rounded corners=1pt,
    minimum width=\featBlockW,
    minimum height=\featBlockH,
    inner sep=0pt}
]


\newcommand{\slabEdgeLW}{0.7pt}

\tikzset{
  nonarrowline/.style={
    draw=darkgray,
    line width=\slabEdgeLW
  },
  redguideline/.style={
    draw=red!70!black,
    line width=\slabEdgeLW
  },
  featuredashbox/.style={
    draw=darkgray,
    dashed,
    dash pattern=on 3pt off 2pt,
    rounded corners=3pt,
    line width=\slabEdgeLW
  }
}

  \node[inner sep=0pt] (person) at (0,0)
    {\includegraphics[width=2.7 cm, trim=5 5 5 5, clip]{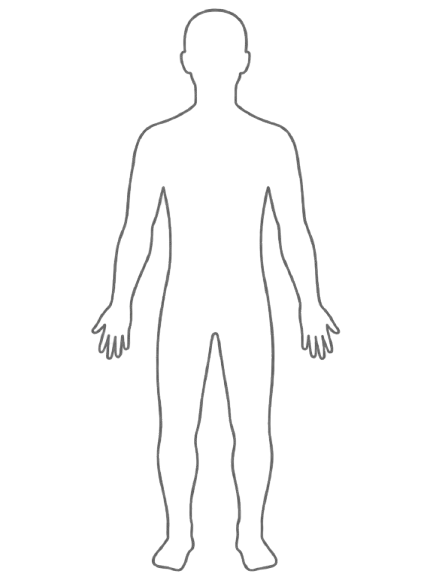}};

\node[inner sep=0pt] (mricube) at ($(person.center)+(0,0.20)$)
  {\includegraphics[width=0.42cm]{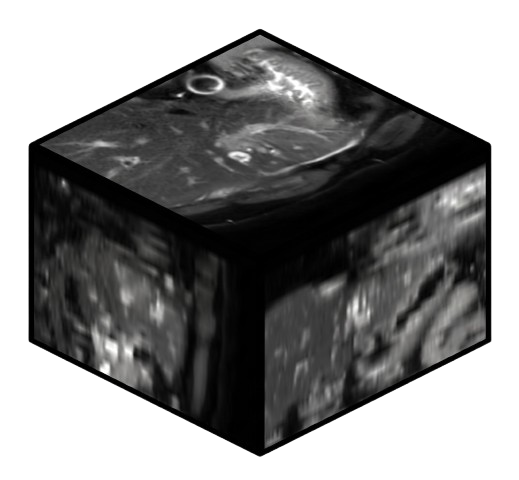}};  

  \node[figlabel, align=center] at (0,-2.225)
    {Tumor-Centered\\MRI Crop};
    

  \node[inner sep=0pt] (sBottom) at (2.45,-1.45)
    {\includegraphics[width=\sliceW]{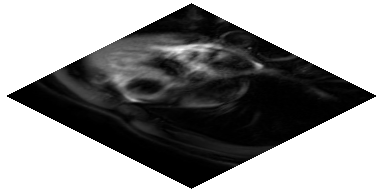}};
  \node[figlabel] at (2.45,-0.65 ) {$\vdots$};

  \node[inner sep=0pt] (selSliceImg) at (2.45,0.2)
  {\includegraphics[width=1.5cm, trim=100 100 100 100, clip]{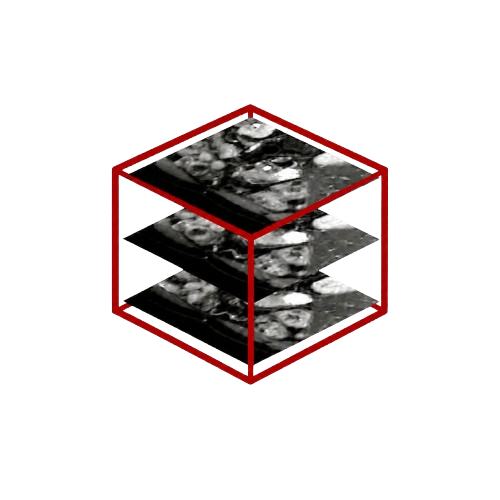}};
  
  \node[
    inner sep=0pt,
    fit=(selSliceImg)
  ] (selSliceGroup) {};

  \draw[
    red!75!black,
    line width=0.55pt,
    decorate,
    decoration={brace, amplitude=3pt,mirror}
  ]
    ($(selSliceImg.west)+(-0.01,0.32)$) --
    ($(selSliceImg.west)+(-0.01,-0.32)$)
    node[midway, left=3pt, red!75!black, font=\scriptsize] {$k$};

  \node[inner sep=0pt] (sTop) at (2.45,0.95)
    {\includegraphics[width=\sliceW]{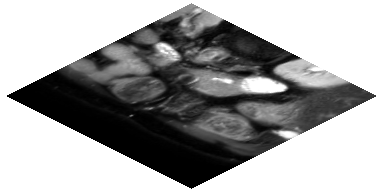}};
    
  \node[figlabel, align=center] at (2.45,-2.225)
    {Axial Slice\\Sequence};

  \coordinate (cropLineTop) at ($(person.center)+(0.10,0.45)$);
  \coordinate (cropLineBot) at ($(person.center)+(0.10,-0.05)$);

  \draw[nonarrowline]
    (cropLineTop) -- (sTop.west);

  \draw[nonarrowline]
    (cropLineBot) -- (sBottom.west);
  

  \node[inner sep=0pt] (bagBottomImg) at (4.75,-1.45)
    {\includegraphics[width=\slabBagW]{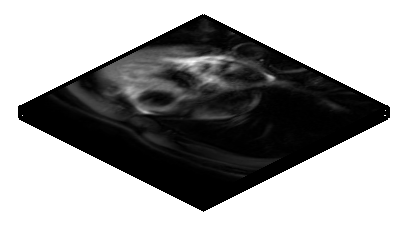}};
  \node[figlabel] at (4.75,-0.65) {$\vdots$};

  \node[inner sep=0pt] (bag3Img) at (4.75,-0.04)
    {\includegraphics[width=\slabBagW]{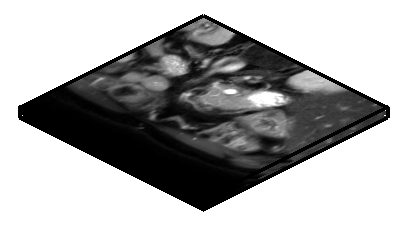}};

  \node[inner sep=0pt] (bag2Img) at (4.75,0.29)
    {\includegraphics[width=\slabBagW]{figures/figure_1/slab_bags/black/slab_z002_004.png}};
  
  \node[inner sep=0pt] (bagSelImg) at (4.75,0.62)
    {\includegraphics[width=\slabBagW]{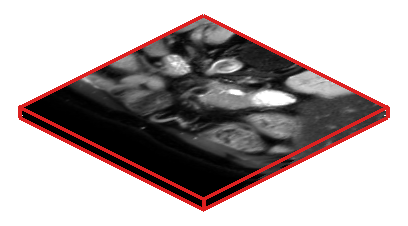}};

  \node[inner sep=0pt] (bag1Img) at (4.75,0.95)
    {\includegraphics[width=\slabBagW]{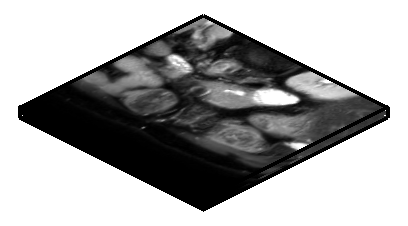}};

  \coordinate (selGuideTop) at ($(selSliceImg.center)+(0.7, 0.40)$);
  \coordinate (selGuideBot) at ($(selSliceImg.center)+(0.7,-0.30)$);

  \coordinate (bagGuideTop) at ($(bagSelImg.center)+(-0.70, 0.03)$);
  \coordinate (bagGuideBot) at ($(bagSelImg.center)+(-0.15,-0.4)$);

  \draw[redguideline]
    (selGuideTop) -- (bagGuideTop);

  \draw[redguideline]
    (selGuideBot) -- (bagGuideBot);

  \node[figlabel, align=center] at (4.75,-2.225)
    {2.5D Slab\\Sequence};

  \node[
    inner sep=0pt,
    fit=(bag1Img)(bag2Img)(bag3Img)(bagSelImg)(bagBottomImg)
  ] (slabs) {};
  
\node[
  opblock,
  fill=silver,
  minimum width=\opBlockH,
  minimum height=\opBlockW,
  right=\encSideGap of slabs
] (enc)
{\rotatebox{90}{\shortstack{Shared Slab\\Encoder}}};


\coordinate (encIn1)   at (enc.west |- bag1Img.east);
\coordinate (encIn2)   at (enc.west |- bag2Img.east);
\coordinate (encIn3)   at (enc.west |- bag3Img.east);
\coordinate (encInSel) at (enc.west |- bagSelImg.east);
\coordinate (encInBot) at (enc.west |- bagBottomImg.east);

\coordinate (encOut1)   at (enc.east |- bag1Img.east);
\coordinate (encOut2)   at (enc.east |- bag2Img.east);
\coordinate (encOut3)   at (enc.east |- bag3Img.east);
\coordinate (encOutSel) at (enc.east |- bagSelImg.east);
\coordinate (encOutBot) at (enc.east |- bagBottomImg.east);


\node[
  draw=gray!70,
  fill=silver,
  rounded corners=1pt,
  minimum width=0.45cm,
  minimum height=0.22cm
] (feat1) at ($(encOut1)+(\featXOffset,0)$) {};

\node[
  draw=gray!70,
  fill=silver,
  rounded corners=1pt,
  minimum width=0.45cm,
  minimum height=0.22cm
] (feat2) at ($(encOut2)+(\featXOffset,0)$) {};

\node[
  draw=gray!70,
  fill=silver,
  rounded corners=1pt,
  minimum width=0.45cm,
  minimum height=0.22cm
] (feat3) at ($(encOut3)+(\featXOffset,0)$) {};

\node[
  draw=gray!70,
  fill=silver,
  rounded corners=1pt,
  minimum width=0.45cm,
  minimum height=0.22cm
] (featSel) at ($(encOutSel)+(\featXOffset,0)$) {};

\node[
  draw=gray!70,
  fill=silver,
  rounded corners=1pt,
  minimum width=0.45cm,
  minimum height=0.22cm
] (featBot) at ($(encOutBot)+(\featXOffset,0)$) {};

\coordinate (featDotsBase) at ($(featSel)!0.5!(featBot)$);
\node[figlabel] (featDots) at ($(featDotsBase)+(0,-0.3)$) {$\vdots$};

\node[
  featuredashbox,
  inner sep=0pt,
  minimum width=\featBoxW,
  minimum height=\opBlockW
] (featBox) at (feat1.center |- enc.center) {};

  \node[figlabel, align=center] at (8.8,-2.225)
    {Slab Feature\\Sequence};
\coordinate (featArrowEnd1)   at ($(feat1.west)+(-0.1,0)$);
\coordinate (featArrowEnd2)   at ($(feat2.west)+(-0.1,0)$);
\coordinate (featArrowEnd3)   at ($(feat3.west)+(-0.1,0)$);
\coordinate (featArrowEndSel) at ($(featSel.west)+(-0.1,0)$);
\coordinate (bottomArrowEnd)  at ($(featBot.west)+(-0.1,0)$);
\begin{scope}[on background layer]
  \draw[arrow] (bag1Img.east)      -- (featArrowEnd1);
  \draw[arrow] (bag2Img.east)      -- (featArrowEnd2);
  \draw[arrow] (bag3Img.east)      -- (featArrowEnd3);
  \draw[arrow] (bagSelImg.east)    -- (featArrowEndSel);
  \draw[arrow] (bagBottomImg.east) -- (bottomArrowEnd);
\end{scope}


\node[figset] (set) at ($(featBox.east)+(\branchXOffset,0.72)$)
{Set-Attention\\Branch};

\node[figord] (ord) at ($(featBox.east)+(\branchXOffset,-0.72)$)
{Sequence-Attention\\Branch};

\draw[arrow]
  (featBox.east |- set.west) -- (set.west);

\draw[arrow]
  (featBox.east |- ord.west) -- (ord.west);

\node[fighead] (head) at ($(set.east)!0.5!(ord.east)+(\headXOffset,0)$)
{\rotatebox{90}{\shortstack{Prediction\\Head}}};

\node[figlabel, align=center, right=\flowGap of head] (pni)
{PNI\\Probability};

\draw[arrow] (set.east) -- (head.west |- set.east);
\draw[arrow] (ord.east) -- (head.west |- ord.east);
\draw[arrow] (head.east) -- (pni.west);

\end{tikzpicture}}}
\caption{Overview of \method{}. A tumor-centered T2-weighted MRI crop is represented as an ordered bag of contiguous 2.5D slabs. A shared encoder extracts slab features. Patient-level probability estimation combines order-agnostic set-attention pooling with sequence-aware attention pooling.}
\label{fig:overview}
\end{figure*}

\section{Methods}
\subsection{Problem Formulation and Preprocessing}
For patient $i\in\{1,\ldots,N\}$, let $X_i\in\mathbb{R}^{H\times W\times D}$ denote the preprocessed tumor-centered T2-weighted MRI crop and let $y_i\in\{0,1\}$ denote the PNI label. The classifier estimates a patient-level logit and probability,
\begin{equation}
\ell_i=f_{\Theta}(X_i), \qquad \hat{p}_i=\sigmoid(\ell_i).
\label{eq:predictor}
\end{equation}
Each crop is resampled to $1.0\times1.0\times5.0\,\mathrm{mm}$ spacing and fixed to a physical field of view of $192\times192\times160\,\mathrm{mm}$, yielding $H=W=192$ and $D=32$. Intensities are clipped at the 0.5th and 99.5th percentiles and normalized within the crop using robust $z$-score normalization.

\subsection{Ordered 2.5D Slab Construction}
Let $\xv_{i,t}\in\mathbb{R}^{H\times W}$ be the axial slice at position $t\in\{1,\ldots,D\}$. For an odd slab size $k=2r+1$, the slab centered at $t$ is
\begin{equation}
S_{i,t}^{(k)}=
\left[\xv_{i,\kappa(t-r)},\ldots,\xv_{i,\kappa(t)},\ldots,\xv_{i,\kappa(t+r)}\right]
\in\mathbb{R}^{H\times W\times k},
\label{eq:slab}
\end{equation}
where $\kappa(u)=\min\{\max(u,1),D\}$ clamps out-of-range indices. This boundary handling keeps one slab per axial position. The ordered patient bag is
\begin{equation}
B_i^{(k)}=\left(S_{i,1}^{(k)},S_{i,2}^{(k)},\ldots,S_{i,D}^{(k)}\right),
\label{eq:bag}
\end{equation}
with the same through-plane order as the original crop. Unless otherwise specified, experiments use $k=3$.

\subsection{Shared Slab Encoder}
Each slab is encoded by a shared ResNet-18 backbone \cite{he2016resnet}. Let $g_{\theta}$ denote the adapted backbone. The slab feature is
\begin{equation}
\zv_{i,t}=g_{\theta}\!\left(S_{i,t}^{(k)}\right)\in\mathbb{R}^{d},
\label{eq:encoder}
\end{equation}
where $d=512$. The same encoder is applied to all slab positions.

The first convolution of the pretrained ResNet-18 is adapted to the slab size using grayscale-based symmetric initialization. If $\W_{\mathrm{rgb}}\in\mathbb{R}^{C_o\times3\times h\times w}$ denotes the pretrained RGB convolution, a grayscale basis is obtained by
\begin{equation}
\bar{\W}=\frac{1}{3}\sum_{a=1}^{3}\W_{\mathrm{rgb}}[:,a,:,:]
\in\mathbb{R}^{C_o\times h\times w}.
\label{eq:gray_basis}
\end{equation}
For a $k$-channel slab input, each channel of the adapted convolution is initialized as
\begin{equation}
\W^{(k)}[:,j,:,:]=\frac{3}{k}\bar{\W},
\qquad j=1,\ldots,k.
\label{eq:slab_conv_init}
\end{equation}
This provides a symmetric channel initialization while keeping the first-layer response scale comparable across slab sizes.

\subsection{Set-Attention Branch}
The set-attention branch performs additive attention pooling over the slab bag. It computes scalar scores and normalizes them across instances:
\begin{align}
 r_{i,t}^{\mathrm{set}}
 &= \vv_{\mathrm{set}}^{\top}\tanh\!\left(\W_{\mathrm{set}}\zv_{i,t}+\bv_{\mathrm{set}}\right),
\label{eq:set_score}\\
 \alpha_{i,t}
 &= \frac{\exp(r_{i,t}^{\mathrm{set}})}{\sum_{u=1}^{D}\exp(r_{i,u}^{\mathrm{set}})}.
\label{eq:set_softmax}
\end{align}
The set-attention summary is
\begin{equation}
\hv_i^{\mathrm{set}}=\sum_{t=1}^{D}\alpha_{i,t}\zv_{i,t}.
\label{eq:set_summary}
\end{equation}
This branch provides a permutation-invariant pooling path.

\subsection{Sequence-Attention Branch}
The sequence-attention branch models the slab features as an axial sequence. A one-layer bidirectional gated recurrent unit (GRU) produces contextual states \cite{cho2014gru,schuster1997bidirectional}:
\begin{align}
\overrightarrow{\qv}_{i,t}
&=\operatorname{GRU}_{\mathrm{f}}(\zv_{i,t},\overrightarrow{\qv}_{i,t-1}),
\label{eq:forward_gru}\\
\overleftarrow{\qv}_{i,t}
&=\operatorname{GRU}_{\mathrm{b}}(\zv_{i,t},\overleftarrow{\qv}_{i,t+1}).
\label{eq:backward_gru}
\end{align}
The two directional states are concatenated and projected to the common feature dimension:
\begin{align}
\qv_{i,t} &= \left[\overrightarrow{\qv}_{i,t};\overleftarrow{\qv}_{i,t}\right],
\label{eq:seq_state_concat}\\
\uv_{i,t} &= \W_{\mathrm{proj}}\qv_{i,t}+\bv_{\mathrm{proj}} \in \mathbb{R}^{d}.
\label{eq:seq_state_projection}
\end{align}
Attention scores and normalized through-plane weights are computed from the projected states:
\begin{align}
 r_{i,t}^{\mathrm{seq}}
 &= \vv_{\mathrm{seq}}^{\top}\tanh\!\left(\W_{\mathrm{seq}}\uv_{i,t}+\bv_{\mathrm{seq}}\right),
\label{eq:seq_score}\\
 \beta_{i,t}
 &= \frac{\exp(r_{i,t}^{\mathrm{seq}})}{\sum_{u=1}^{D}\exp(r_{i,u}^{\mathrm{seq}})}.
\label{eq:seq_softmax}
\end{align}
The sequence-attention summary is
\begin{equation}
\hv_i^{\mathrm{seq}}=\sum_{t=1}^{D}\beta_{i,t}\uv_{i,t}.
\label{eq:seq_summary}
\end{equation}

\definecolor{jjy_white}{RGB}{255,255,255}
\definecolor{jjy_gray}{RGB}{192,192,192}
\definecolor{jjy_apricot}{RGB}{255,238,218}
\definecolor{jjy_coral}{RGB}{255,204,190}

\colorlet{jjy_orange}{jjy_gray}  
\colorlet{jjy_pink}{jjy_gray}    
\colorlet{jjy_red}{jjy_white}    
\colorlet{jjy_green}{jjy_apricot}
\colorlet{jjy_purple}{jjy_gray}  
\colorlet{jjy_skyblue}{jjy_gray} 
\colorlet{jjy_peach}{jjy_white}  
\colorlet{jjy_amber}{jjy_white}  
\colorlet{jjy_mint}{jjy_coral}   
\colorlet{jjy_blue}{jjy_coral}   

\tikzset{
  jjy_arrow/.style={-Latex, thick, draw=black!75},
  jjy_line/.style={thick, draw=black!75},
  jjy_block/.style={draw=black, line width=0.6pt, rounded corners=3pt, align=center, text=black},
  jjy_op/.style={jjy_block, minimum width=3.0cm, minimum height=1.0cm, font=\normalsize},
  jjy_gru/.style={jjy_block, minimum width=1.0cm, minimum height=0.50cm, rounded corners=2pt, font=\normalsize},
  jjy_state/.style={draw=black, line width=0.6pt, circle, minimum size=0.8cm, inner sep=0pt, align=center, font=\normalsize}
}

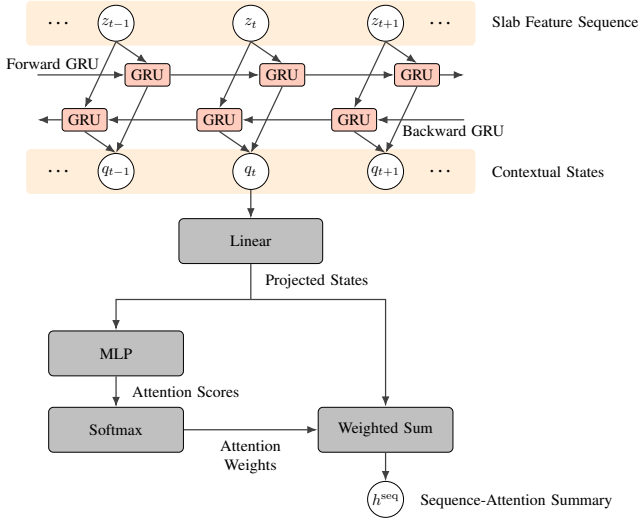
\begin{figure}[t]
  \centering
  \makebox[\columnwidth][c]{%
  \resizebox{0.98\columnwidth}{!}{%
    \begin{tikzpicture}[scale=1]
      \fill[jjy_green, rounded corners=3pt] (-2, 0.4) rectangle (8, -0.6);
      \node[text width=100pt, anchor=west, align=left, font=\normalsize, text=black] at (8.25, -0.1) {Slab Feature Sequence};

      \fill[jjy_green, rounded corners=3pt] (-2, -2.9) rectangle (8, -3.9);
      \node[text width=75pt, anchor=west, align=left, font=\normalsize, text=black] at (8.25, -3.4) {Contextual States};

      \node at (-1.25,-0.1) {$\bm{\cdots}$};
      \node[jjy_state, fill=jjy_peach] (zprev) at (0,-0.1) {$z_{t-1}$};
      \node[jjy_state, fill=jjy_peach] (zt)    at (3,-0.1) {$z_{t}$};
      \node[jjy_state, fill=jjy_peach] (znext) at (6,-0.1) {$z_{t+1}$};
      \node at (7.25,-0.1) {$\bm{\cdots}$};

      \node[text width=60pt, anchor=east, align=right, font=\normalsize, text=black] at (-0.25, -1) {Forward GRU};
      \node[text width=70pt, anchor=west, align=left, font=\normalsize, text=black] at (6.25, -2.5) {Backward GRU};

      \node[jjy_gru, fill=jjy_mint] (fgru1) at (0.7,-1.25) {GRU};
      \node[jjy_gru, fill=jjy_mint] (fgru2) at (3.7,-1.25) {GRU};
      \node[jjy_gru, fill=jjy_mint] (fgru3) at (6.7,-1.25) {GRU};

      \node[jjy_gru, fill=jjy_blue] (bgru1) at (-0.7,-2.25) {GRU};
      \node[jjy_gru, fill=jjy_blue] (bgru2) at (2.3,-2.25) {GRU};
      \node[jjy_gru, fill=jjy_blue] (bgru3) at (5.3,-2.25) {GRU};

      \node at (-1.25,-3.4) {$\bm{\cdots}$};
      \node[jjy_state, fill=jjy_amber] (qprev) at (0,-3.4) {$q_{t-1}$};
      \node[jjy_state, fill=jjy_amber] (qt)    at (3,-3.4) {$q_{t}$};
      \node[jjy_state, fill=jjy_amber] (qnext) at (6,-3.4) {$q_{t+1}$};
      \node at (7.25,-3.4) {$\bm{\cdots}$};

      \draw[jjy_arrow] (-1.75,-1.25) -- (fgru1.west);
      \draw[jjy_arrow] (fgru1.east) -- (fgru2.west);
      \draw[jjy_arrow] (fgru2.east) -- (fgru3.west);
      \draw[jjy_arrow] (fgru3.east) -- (7.75,-1.25);

      \draw[jjy_arrow] (bgru1.west) -- (-1.75,-2.25);
      \draw[jjy_arrow] (bgru2.west) -- (bgru1.east);
      \draw[jjy_arrow] (bgru3.west) -- (bgru2.east);
      \draw[jjy_arrow] (7.75,-2.25) -- (bgru3.east);

      \draw[jjy_arrow] (zprev.south) -- (fgru1.north);
      \draw[jjy_arrow] (zt.south)    -- (fgru2.north);
      \draw[jjy_arrow] (znext.south) -- (fgru3.north);

      \draw[jjy_arrow] (zprev.south) -- (bgru1.north);
      \draw[jjy_arrow] (zt.south)    -- (bgru2.north);
      \draw[jjy_arrow] (znext.south) -- (bgru3.north);

      \draw[jjy_arrow] (bgru1.south) -- (qprev.north);
      \draw[jjy_arrow] (bgru2.south) -- (qt.north);
      \draw[jjy_arrow] (bgru3.south) -- (qnext.north);

      \draw[jjy_arrow] (fgru1.south) -- (qprev.north);
      \draw[jjy_arrow] (fgru2.south) -- (qt.north);
      \draw[jjy_arrow] (fgru3.south) -- (qnext.north);

      \draw[jjy_arrow] (qt.south) -- (3, -4.45);
      \node[jjy_op, fill=jjy_purple, minimum width=3.2cm] (linear) at (3,-4.95) {Linear};

      \coordinate (projSplit) at (3, -6.25);
      \draw[jjy_line] (linear.south) -- (projSplit);
      \node[anchor=west, align=left, font=\normalsize, text=black] at (3.20, -5.85) {Projected States};

      \node[jjy_op, fill=jjy_orange] (mlp) at (0,-7.45) {MLP};
      \draw[jjy_arrow] (projSplit) -- (0, -6.25) -- (mlp.north);

      \node[jjy_op, fill=jjy_skyblue] (wsum) at (6, -9.15) {Weighted Sum};
      \draw[jjy_arrow] (projSplit) -- (6, -6.25) -- (wsum.north);

      \draw[jjy_arrow] (mlp.south) -- (0, -8.65);
      \node[text width=150pt, anchor=west, align=left, font=\normalsize, text=black] at (0.25, -8.30) {Attention Scores};

      \node[jjy_op, fill=jjy_pink] (softmax) at (0,-9.15) {Softmax};
      \draw[jjy_arrow] (softmax.east) -- (wsum.west);
      \node[text width=150pt, anchor=north, align=center, font=\normalsize, text=black] at (3, -9.25) {Attention\\Weights};

      \draw[jjy_arrow] (wsum.south) -- (6, -10.30);
      \node[jjy_state, fill=jjy_red] (hseq) at (6,-10.70) {$h^{\mathrm{seq}}$};
      \node[text width=125pt, anchor=west, align=left, font=\normalsize, text=black] at (6.65, -10.75) {Sequence-Attention Summary};
    \end{tikzpicture}%
  }}
  \caption{Sequence-attention branch. Bidirectional recurrent context is projected to the common feature dimension before attention scoring; aggregation weights and the weighted summary are computed from the projected sequence states.}
  \label{fig:ordered_branch}
\end{figure}

\subsection{Prediction Head and Training Objective}
The branch summaries are normalized with a shared LayerNorm, concatenated, and passed through a prediction head:
\begin{align}
\hv_i^{\mathrm{cat}}
&= [\LN(\hv_i^{\mathrm{set}});\LN(\hv_i^{\mathrm{seq}})],
\label{eq:prediction_concat}\\
\hv_i^{\mathrm{pred}}
&= \GELU(\W_1\hv_i^{\mathrm{cat}}+\bv_1),
\label{eq:prediction_hidden}\\
\ell_i
&= \vv_2^{\top}\hv_i^{\mathrm{pred}}+b_2.
\label{eq:prediction_logit}
\end{align}
Dropout is applied during training in the feed-forward components. The network is optimized with positive-class-weighted binary cross-entropy with logits, where the positive-class weight is computed from the training split.

\section{Experiments and Results}
\subsection{Data and Evaluation Protocol}
The evaluation used a single-center cohort of 183 patients who underwent preoperative T2-weighted MRI, including 70 PNI-positive and 113 PNI-negative patients. Tumor-centered crops were extracted from the MRI based on expert-annotated tumor masks created with 3D Slicer \cite{fedorov2012slicer}. 
Performance was estimated using five-fold label-stratified cross-validation at the patient level. In each split, three folds were used for training, one fold was used for checkpoint selection and early stopping, and the remaining fold was held out for testing. Volumetric baselines operated on the preprocessed crop, whereas MIL baselines operated on the same ordered 2.5D slab instances. Unless otherwise specified, performance values are reported as mean $\pm$ standard deviation across the five held-out test folds.


AUROC was the primary metric, and AUPRC was used as a complementary metric for the imbalanced endpoint \cite{hanley1982meaning,steyerberg2010assessing,saito2015precision}. DeLong's method was used for the paired AUROC comparison between \method{} and CLAM-SB \cite{delong1988comparing}.

\subsection{Implementation Details}
The model was optimized end-to-end using AdamW. The encoder learning rate was $3\times10^{-5}$, and the aggregation and prediction heads used $3\times10^{-4}$. Training used a batch size of 4 for at most 50 epochs, with early stopping patience 10 based on validation AUROC. Dropout was set to 0.25. The sequence reader was a one-layer bidirectional GRU with hidden size 128 per direction. The attention dimension was 128 for both the set-attention and sequence-attention branches. The weighted binary cross-entropy objective was implemented with a numerically stable logits-based loss, and automatic mixed precision was used during training. Experiments were run on an NVIDIA RTX PRO 6000 Blackwell Server Edition GPU.

\begin{table}[t]
\centering
\caption{Performance comparison with volumetric and MIL baselines.}
\label{tab:baseline}
\scriptsize
\setlength{\tabcolsep}{3.0pt}
\begin{tabular}{@{}L{0.51\columnwidth}C{0.22\columnwidth}C{0.22\columnwidth}@{}}
\toprule
Model & AUROC & AUPRC \\
\midrule
\multicolumn{3}{@{}l}{\textit{Volumetric models}} \\
ResNet-18 \cite{he2016resnet} & $0.645\pm0.090$ & $0.563\pm0.058$ \\
DenseNet-121 \cite{huang2017densenet} & $0.656\pm0.017$ & $0.585\pm0.066$ \\
Vision Transformer \cite{dosovitskiy2021vit} & $0.573\pm0.045$ & $0.457\pm0.035$ \\
Swin Transformer \cite{tang2022swinunetr} & $0.642\pm0.053$ & $0.531\pm0.054$ \\
\midrule
\multicolumn{3}{@{}l}{\textit{MIL models}} \\
ABMIL \cite{ilse2018attention} & $0.675\pm0.080$ & $0.605\pm0.109$ \\
CLAM-SB \cite{lu2021clam} & $0.705\pm0.071$ & $0.608\pm0.118$ \\
DSMIL \cite{li2021dsmil} & $0.678\pm0.100$ & $0.585\pm0.131$ \\
TransMIL \cite{shao2021transmil} & $0.639\pm0.061$ & $0.549\pm0.095$ \\
\method{} (ours) & $\mathbf{0.770}\pm0.077$ & $\mathbf{0.692}\pm0.093$ \\
\bottomrule
\end{tabular}
\end{table}

\subsection{Comparison with Baselines}
Table~\ref{tab:baseline} compares volumetric models with MIL-based alternatives. \method{} achieved the highest mean performance among the evaluated models, with an AUROC of $0.770\pm0.077$ and an AUPRC of $0.692\pm0.093$. The strongest MIL baseline was CLAM-SB, which achieved an AUROC of $0.705\pm0.071$ and an AUPRC of $0.608\pm0.118$. Beyond the fold-wise summaries, DeLong's test showed a positive paired AUROC difference for \method{} over CLAM-SB, with a 95\% confidence interval of $[0.001,\,0.169]$.

\begin{table}[t]
\centering
\caption{Ablation analysis under the fixed end-to-end training protocol.}
\label{tab:ablation}
\scriptsize
\setlength{\tabcolsep}{2.6pt}
\begin{tabular}{@{}L{0.51\columnwidth}C{0.22\columnwidth}C{0.22\columnwidth}@{}}
\toprule
Variant & AUROC & AUPRC \\
\midrule
\multicolumn{3}{@{}l}{\textit{Instance representation}} \\
Single-slice instances ($k=1$) & $0.757\pm0.058$ & $0.665\pm0.071$ \\
3-slice 2.5D slabs ($k=3$, ours) & $\mathbf{0.770}\pm0.077$ & $\mathbf{0.692}\pm0.093$ \\
5-slice 2.5D slabs ($k=5$) & $0.727\pm0.052$ & $0.665\pm0.077$ \\
\midrule
\multicolumn{3}{@{}l}{\textit{Branch aggregation}} \\
Set-attention branch only & $0.703\pm0.046$ & $0.632\pm0.078$ \\
Sequence-attention branch only & $0.725\pm0.080$ & $0.673\pm0.128$ \\
Set- and sequence-attention branches (ours) & $\mathbf{0.770}\pm0.077$ & $\mathbf{0.692}\pm0.093$ \\
\midrule
\multicolumn{3}{@{}l}{\textit{Sequence reader}} \\
1D convolution & $0.706\pm0.069$ & $0.626\pm0.068$ \\
Transformer encoder & $0.738\pm0.101$ & $0.691\pm0.120$ \\
Unidirectional GRU & $0.724\pm0.061$ & $0.655\pm0.080$ \\
Bidirectional GRU (ours) & $\mathbf{0.770}\pm0.077$ & $\mathbf{0.692}\pm0.093$ \\
Bidirectional LSTM & $0.736\pm0.071$ & $0.644\pm0.056$ \\
\bottomrule
\end{tabular}
\end{table}

\subsection{Ablation Analysis}
Table~\ref{tab:ablation} evaluates the slab representation, branch aggregation, and sequence reader under the same training protocol.
The ablation results indicate that the 3-slice slab representation, fused set- and sequence-attention aggregation, and bidirectional GRU reader produced the strongest overall results under the fixed training protocol.

\subsection{Crop Localization Sensitivity}
To assess sensitivity to crop localization, we performed a test-time perturbation analysis without retraining. For each case, we generated three perturbed crops by independently sampling in-plane offsets from $\mathcal{U}(-10,10)\,\mathrm{mm}$ and a through-plane offset from $\{-1,0,+1\}$ axial slices. With perturbed crops, \method{} achieved an AUROC of $0.759\pm0.069$ and an AUPRC of $0.682\pm0.052$, remaining close to the unperturbed performance. These results indicate that performance was largely preserved within the tested perturbation range.

\subsection{Computational Profile}
Table~\ref{tab:compute} reports model size, multiply--accumulate operations, peak GPU memory, and forward time. Peak memory denotes the maximum allocated GPU memory during one forward pass after resetting peak-memory statistics. Forward time is the median latency over 10 forward passes after 5 warm-up passes on a single GPU.

\begin{table}[t]
\centering
\caption{Computational profiles of the evaluated models.}
\label{tab:compute}
\scriptsize
\setlength{\tabcolsep}{1.4pt}
\renewcommand{\arraystretch}{1.05}
\begin{tabular}{@{}L{0.35\columnwidth}C{0.15\columnwidth}C{0.15\columnwidth}C{0.15\columnwidth}C{0.15\columnwidth}@{}}
\toprule
Model &
\shortstack{Params.\\(M)} &
\shortstack{MACs\\(G)} &
\shortstack{Peak mem.\\(MB)} &
\shortstack{Time\\(ms)} \\
\midrule
\multicolumn{5}{@{}l}{\textit{Volumetric models}} \\
ResNet-18 & 33.15 & 295.64 & 709.05 & 12.19 \\
DenseNet-121 & 25.22 & 51.87 & 242.35 & 9.11 \\
Vision Transformer & 51.89 & 152.81 & 629.45 & 16.82 \\
Swin Transformer & 15.70 & 13.91 & 2124.71 & 18.41 \\
\midrule
\multicolumn{5}{@{}l}{\textit{MIL models}} \\
ABMIL & 11.47 & 42.65 & 235.45 & 3.00 \\
CLAM-SB & 11.47 & 42.65 & 235.45 & 2.48 \\
DSMIL & 11.57 & 42.65 & 235.82 & 2.82 \\
TransMIL & 14.10 & 42.73 & 245.48 & 8.24 \\
\method{} (ours) & 13.25 & 42.66 & 208.72 & 3.45 \\
\bottomrule
\end{tabular}
\end{table}

\subsection{Qualitative Analysis}
\begin{figure}[t]
\centering
\includegraphics[width=0.96\linewidth]{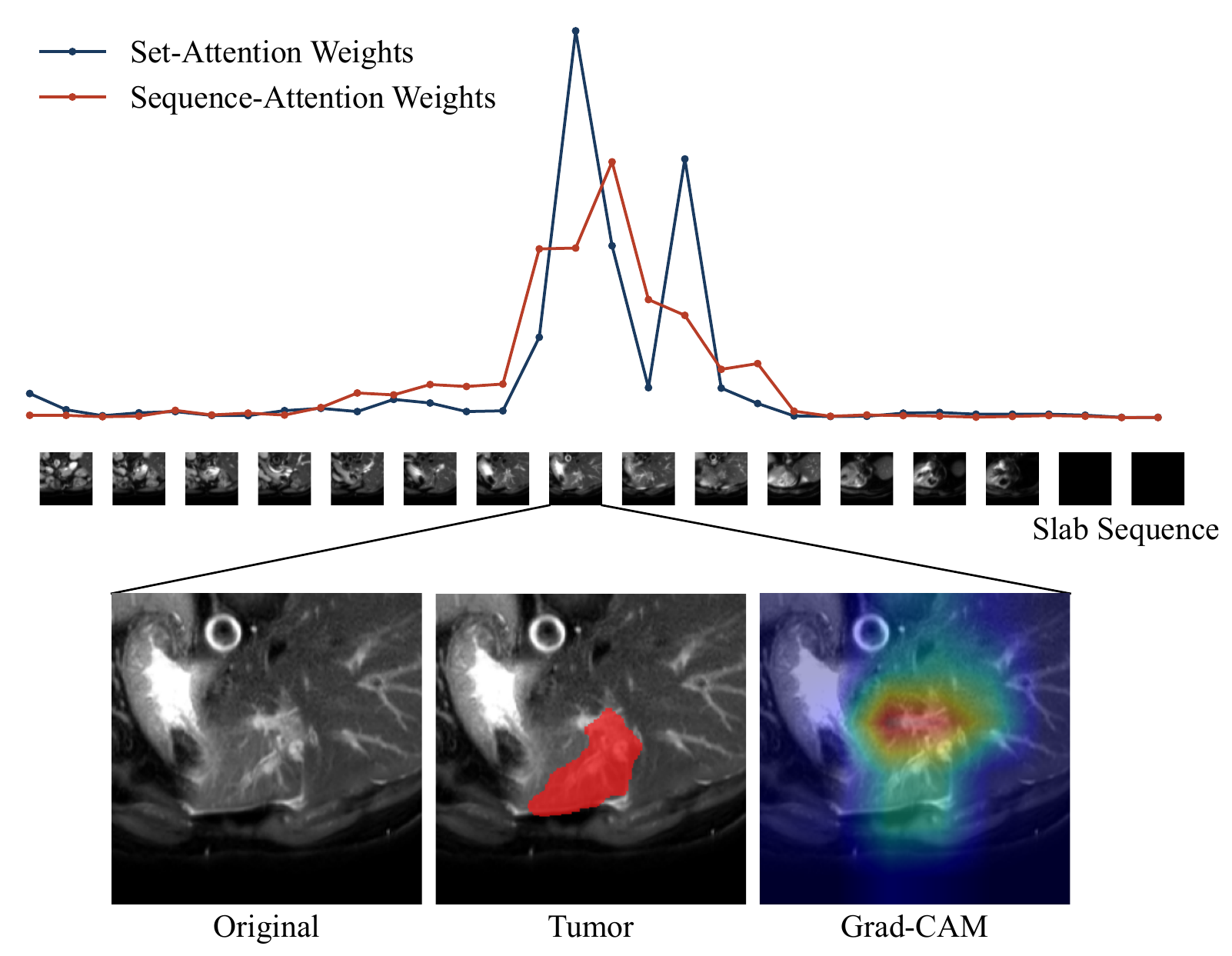}
\caption{Qualitative analysis of a PNI-positive patient. The top panel shows the branch-wise aggregation-weight profiles over the ordered 2.5D slab montage. The bottom panel shows the selected slab, the tumor-mask overlay used for visual reference, and the Grad-CAM overlay.}
\label{fig:qualitative}
\end{figure}

For qualitative analysis, we visualize the set-attention and sequence-attention weights on the ordered slab sequence and compute Grad-CAM from the patient-level logit \cite{selvaraju2017gradcam}. Figure~\ref{fig:qualitative} shows a PNI-positive example. In this example, the aggregation weights concentrate over a limited slab interval near the tumor-containing region, and Grad-CAM shows a coarse response around the tumor and adjacent tissue. These outputs are qualitative audit cues rather than voxel-level PNI localization.

\section{Discussion}



\method{} achieved the highest mean discrimination among the evaluated models, suggesting that sequence-aware aggregation can complement permutation-invariant MIL pooling for patient-level PNI prediction. The ablation results further showed that the 3-slice slab representation, combined set- and sequence-attention branches, and bidirectional GRU provided the strongest overall performance. The main limitation is the small single-center cohort, which limits the precision and generalizability of the performance estimates. External validation is needed to assess robustness across scanners, imaging protocols, and institutions.

\section{Conclusion}
This paper presented \method{}, an order-aware 2.5D MIL framework for predicting PNI from preoperative T2-weighted MRI in patients with ICC. The method represents each tumor-centered MRI crop as an ordered slab bag and combines set-attention pooling with sequence-aware aggregation. Using five-fold label-stratified cross-validation at the patient level, \method{} achieved a mean AUROC of $0.770 \pm 0.077$, outperforming the evaluated volumetric and MIL baselines. These results suggest that axial order provides a useful inductive bias for weakly supervised PNI prediction from MRI in this cohort.


\section*{Acknowledgment} 
This work was supported by a grant from the Institute of Information \& Communications Technology Planning \& Evaluation (IITP), funded by the Ministry of Science and ICT (MSIT), Republic of Korea (No. 2020-0-01305, ``Development of AI Deep-Learning Processor and Module for 2,000 TFLOPS Server'').

\bibliographystyle{IEEEtran}
\bibliography{refs}

\end{document}